\documentclass[]{beingbeyond}
\usepackage{enumitem}
\usepackage[toc,page,header]{appendix}

\usepackage{url}
\usepackage[utf8]{inputenc} 
\usepackage[T1]{fontenc}    
\usepackage{url}            
\usepackage{array}          
\usepackage{booktabs}       
\usepackage{amsfonts}       
\usepackage{nicefrac}       
\usepackage{microtype}      
\usepackage{xcolor}         
\usepackage{xspace}
\usepackage{bm}
\usepackage{bbm}
\usepackage{tabularx}
\usepackage{amssymb}
\usepackage{enumitem}
\usepackage{amsmath}
\usepackage{mathtools}
\usepackage{amsthm}
\usepackage{multirow}
\usepackage{makecell}
\usepackage{color}
\usepackage{colortbl}
\usepackage{adjustbox}
\usepackage{caption}
\usepackage{graphicx}
\usepackage{wrapfig}
\usepackage{tcolorbox}
\usepackage{subcaption}
\usepackage{bbding}
\usepackage{pifont}
\usepackage{textcomp}

\newcommand{\DataName}{\texttt{OpenT2M}\xspace}
\newcommand{\ModelName}{\texttt{MonoFrill}\xspace}

\tcbset{
    mybox/.style={
        colback=gray!20, 
        colframe=black,  
        arc=0pt,         
        boxrule=0.5pt,   
        left=6pt,        
        right=6pt,       
        top=6pt,         
        bottom=6pt,      
    }
}

\definecolor{HeaderColor}{RGB}{250, 248, 242} 
\definecolor{BlockA}{RGB}{175, 205, 210}
\definecolor{BlockB}{RGB}{200, 195, 215}

\title{OpenT2M: No-frill Motion Generation with Open-source, Large-scale, High-quality Data}

\author{{\bfseries Bin Cao$^{1,2,3}$ \quad Sipeng Zheng$^{6}$ \quad Hao Luo$^{5}$ \quad Boyuan Li$^{4}$ \\ \quad Jing Liu$^{1,2,\dagger}$ \quad Zongqing Lu$^{5,6,\dagger}$}}

\affiliation{{$^{1}$CASIA 
        \quad $^{2}$UCAS
        \quad $^{3}$BAAI
        \quad $^{4}$RUC
        \quad $^{5}$PKU
        \quad $^{6}$BeingBeyond
        }}

\webpage{\url{https://research.beingbeyond.com/opent2m}}

\firstfig[width=\linewidth][\textwidth]
  {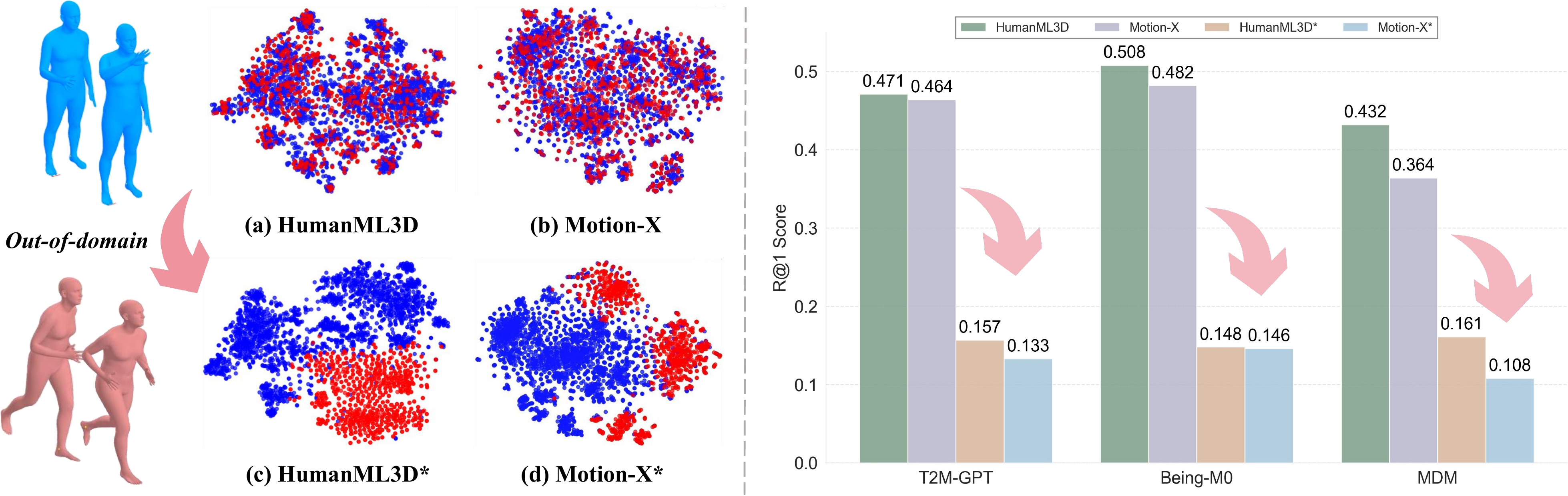}
  {\textbf{(Left)} Visualization of text embeddings for the \textcolor{blue}{training} and \textcolor{red}{validation} sets of HumanML3D and Motion-X. 
        A substantial overlap between the splits indicates data leakage. 
        To avoid this risk, we remove the overlap via data repartition (version denoted as $*$).
        \textbf{(Right)} However, we observe a drastic performance drop when experimenting on this repartitioned benchmark, which reveals the limited generalization capability of current methods when faced with out-of-domain data.}
  {fig:intro}

\abstract{
Text-to-motion (T2M) generation aims to create realistic human movements from text descriptions, with promising applications in animation and robotics.
Despite recent progress, current T2M models perform poorly on unseen text descriptions due to the small scale and limited diversity of existing motion datasets.
To address this problem, we introduce \textbf{\DataName}, a \textbf{million-level}, \textbf{high-quality}, and \textbf{open-source} motion dataset containing over 2800 hours of human motion. 
Each sequence undergoes rigorous quality control through physical feasibility validation and multi-granularity filtering, with detailed second-wise text annotations. 
We also develop an automated pipeline for creating long-horizon sequences, enabling complex motion generation.
Building upon \DataName, we introduce \ModelName, a pretrained motion model that achieves compelling T2M results without complicated designs or technique tricks as ``frills''.
Its core component is 2D-PRQ, a novel motion tokenizer that captures spatiotemporal dependencies by dividing the human body into biology parts.
Experiments show that \DataName significantly improves generalization of existing T2M models, while 2D-PRQ achieves superior reconstruction and strong zero-shot performance.
We expect \DataName and \ModelName will advance the T2M field by addressing longstanding data quality and benchmarking challenges.

}

\checkdata[Date]{March 19, 2026}

\begingroup
\footnotetext{$\ddagger$ Correspondence to \textless jliu@nlpr.ia.ac.cn \textgreater \ and \ \textless zongqing.lu@pku.edu.cn \textgreater}
 
\endgroup

\begin{document}

\maketitle

\section{Introduction}
\label{sec:intro} 
Recent years have seen remarkable progress in generating human motion according to text descriptions for video games, movies, and humanoid robots. 
However, current state-of-the-art methods~\cite{guo2024momask,jiang2023motiongpt}, which depend heavily on motion-capture data~\cite{mahmood2019amass,guo2022generating}, struggle to create novel motions beyond what they've seen during training. 
We argue that this limited generalization in text-to-motion (T2M) models arises from a fundamental bottleneck on existing motion datasets: they lack both diversity and scale.
In fact, we suppose that many reported improvements on standard benchmarks may simply reflect overfitting to the training distribution rather than practical advances. 
To support this claim, we first perform a systematic statistical analysis.

Specifically, we draw the distribution of text descriptions in two widely-used benchmarks: HumanML3D and Motion-X~\cite{lin2023motion} using CLIP~\cite{radford2021learning}.
We observe significant overlap between training and validation sets (Figure \ref{fig:intro}).
Specifically, 10.62\% and 16.97\% validation texts appear word-for-word in the training set in these two datasets --- most of them correspond to quite similar motions.
We also find duplicate descriptions within the validation sets themselves. 
This data contamination seriously undermines how we evaluate T2M models.
To fix this problem, we create their cleaned version (marked with $^*$).
As expected, models perform poorly on the new cleaned benchmarks.
Another concerning issue is that modern T2M methods typically need hundreds of training epochs to converge --- a sign of overfitting, suggesting that existing performance metrics are artificially inflated.

One straightforward thought is to create larger and more diverse motion datasets. 
However, progress in high-quality human motion data has stalled since AMASS was released, mainly because professional motion-capture equipment and facilities are extremely expensive. 
To avoid these costs, recent works~\cite{wang2024quo,fan2025go} have tried extracting motion from internet videos using off-the-shelf estimation tools~\cite{shin2024wham}.
While web videos provide access to diverse motion patterns, this approach brings additional noise.
Most importantly, a large portion of motions extracted from videos contain physically unrealistic artifacts like foot sliding, body drifting, and limb intersections, which severely limit their usefulness for training reliable motion generation models~\cite{holden@2024}.

To solve these problems, we introduce \DataName, a large-scale, high-quality human motion dataset containing over one million sequences. 
Our dataset focuses on bridging the quality gap towards motion-capture databases like HumanML3D while being much larger in scale.
The key advantage of \DataName is that it's freely available to researchers and uses a carefully designed curation process. 
Unlike previous large-scale video-based motion datasets, which are either not publicly available~\cite{wang2024quo} or lack proper physical-aware quality control, we make our dataset open-source with an effective refinement pipeline. 
\DataName offers four key improvements over existing datasets.
\textbf{(1) Physically Feasible Validation:} 
We validate that all motion sequences are physically feasible and can be simulated, making them suitable for training models that control humanoid robots.
\textbf{(2) Multi-granularity Quality Filtering:} 
We remove sequences with occlusions or partial body captures, ensuring that the full human body is visible throughout each motion sequence.
\textbf{(3) Second-wise Descriptions:} 
We generate detailed textual labels for per-second motion, combining them into comprehensive descriptions that accurately capture all actions in the video.
\textbf{(4) Long-horizon Motions:} 
Our dataset includes extended motion sequences that enable models to generate realistic, long-horizon movements from complex text descriptions.
In addition, the increasing scale of motion datasets also poses a challenge for motion tokenizers in accurately reconstructing motions. 
Inspired by residual vector quantization (RQ) techniques~\cite{lee2022autoregressive, guo2024momask} and MotionBook~\cite{wang2024quo}, we propose a novel motion tokenizer, named 2D-PRQ, that shows superior reconstruction performance and great zero-shot ability. Our contributions are summarized as follows:

\begin{itemize}[leftmargin=1.5em]
\item\textbf{A Large-scale, High-quality Motion Database.}
We curate \textbf{\DataName} containing over one million sequences. 
Our dataset ensures that all motions are physically realistic through multi-granularity quality filtering and manual validation. 
It also includes long-horizon motion sequences that enable T2M models to generate complex movements from detailed text descriptions.
\item\textbf{A New Robust Foundation Benchmark.}
In addition to improve the generalization of current T2M models, \DataName further provides a reliable benchmark for fairly evaluating existing methods.
\item\textbf{An Effective No-frill T2M Model.}
We develop a powerful yet ``no-frill'' motion generation model that achieves excellent T2M performance without complicated designs or technical trick.
By simply a novel motion tokenizer named 2D-PRQ, \ModelName effectively captures how motion unfolds both in space and over time. 
After pretraining on \DataName, it shows outstanding performance, especially when tested under zero-shot setups. 
\end{itemize}
\section{Related Work}
\label{sec:related_work}

\noindent\textbf{Human Motion Dataset.} 
Dataset is the foundation of building a robust T2M model. Pioneering datasets, like KIT~\citep{plappert2016kit} and AMASS~\citep{mahmood2019amass}, adapt motion capture devices to obtain human motion data and manual text annotation. The scale and diversity of these datasets are limited. BABEL~\citep{punnakkal2021babel} provides frame-level text annotation on AMASS and serves as a long-horizon motion generation benchmark. HumanML3D~\citep{guo2022generating} expands datasets with 14.6K motions and 44.9K texts by merging AMASS and HumanAct12~\citep{guo2020action2motion}. Motion-X~\citep{lin2024motion} further scales up the dataset by extracting motions from monocular videos and annotating motions by PoseScript~\citep{delmas2022posescript}, resulting in a motion dataset comprising 81.1k sequences. 
~\citet{wang2024quo} introduces the first million-level motion dataset, MotionLib, and highlights the importance of scaling datasets. 
HuMo100M~\citep{cao2025being} is the largest motion dataset featuring 5M motion sequences with multi-granularity text annotation. However, the scarcity of large-scale, high-quality, and open-source datasets hinders building a generalizable T2M model. 
In this work, we introduce \DataName, a large-scale, high-quality, and open-source dataset that improves the generalization ability of current T2M models. 

\noindent\textbf{Motion Tokenization.} Building an effective motion tokenizer is crucial for high-quality motion generation. Motion tokenizer contains a motion encoder, a motion decoder, and a quantizer. T2M-GPT~\citep{zhang2023generating} adapts VQ-VAE to discrete motion into motion tokens by applying the 1D convolution and an embedding to represent the whole body feature. Furthermore, to reduce reconstruction error, ~\citet{lee2022autoregressive} introduces residual quantization (RQ), utilizing multiple layers to quantify motion sequences iteratively. 
Recently, emerging research has explored fine-grained motion tokenization. ~\citet{chen2025language} decouples the human body into the upper body and lower body, and ~\citet{cao2025being} decouples the human body into five parts. However, these methods encode and quantify different body parts independently without skeletal constraints. This limitation motivates us to design 2D-PRQ, a novel motion tokenizer capturing spatial and temporal dependencies and showing superior zero-shot performance. 

\begin{figure*}[t]
\centering
\includegraphics[width=1.0\linewidth]{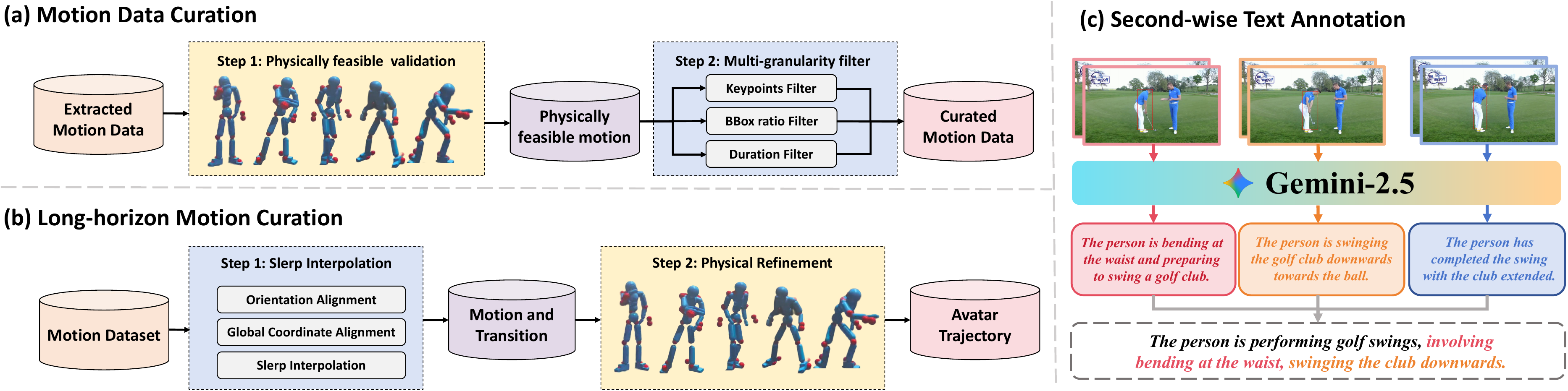}
\caption{\textbf{Data Curation pipeline.} \textbf{(a)} We adopt a two-stage pipeline, including physically feasible validation and multi-granularity filter. \textbf{(b)} We adapt the interpolation-based method for motion curation and introduce an RL-policy for refinement. 
\textbf{(c)} For text annotation, we generate temporally aligned labels for each second of video, using them to synthesize a precise, semantic-rich description.}
\label{fig:data_selection}
\end{figure*}

\section{The OpenT2M Dataset}
The development of robust T2M models is hindered by the lack of large-scale, high-quality data.
Prior datasets suffer from insufficient diversity, often leading to the artifact where R@1 exceeds R@1 Real~\citep{zhang2025motion,tanaka2025unlocking,petrovich2023tmr}, indicating an ambiguous, one-to-many text-motion mapping. 
To address this challenge, we introduce \textbf{\DataName}, an \textbf{open-source} dataset created through a rigorous curation pipeline designed with several key steps (Figure~\ref{fig:data_selection}):

\noindent\textbf{Physically Feasible Validation.}
Motion capture (MoCap) data provides high-quality human motion sequences, valued for its inherent accuracy and adherence to physical constraints ~\citep{mahmood2019amass,guo2025snapmogen}. 
However, MoCap data is difficult to scale up.
To leverage more abundant but noisier video-based motion data, we introduce an RL-based filter to ensure physical plausibility.
We train a robustness policy, $\pi_{\text{refine}}$~\citep{luo2023perpetual} on AMASS, using it to track motions extracted from web videos. 
By retaining only the motions that our policy can successfully track, we eliminate artifacts like jittering and foot-sliding to guarantee physical feasibility.
In our data, more than 63\% of the extracted motions pass this physically feasible validation.
Noting that this process not only optimizes the motion quality, but also maintains highly dynamic motion sequences (e.g, dancing, fencing, and pitching). 
To prove this, we provide additional visualization examples in Appendix~\ref{sec:visualiztion_examples}.
Compared with previous works~\citep{fan2025go,lu2025scamo}, this process ensures the extracted motions adhere to physical constraints, significantly enhancing realism and quality.

\begin{table*}[t]
\centering
\renewcommand{\arraystretch}{0.6}
\setlength{\tabcolsep}{12pt}
\caption{Comparison with existing human motion datasets, where ``\#physically-feasible'' refers to the motion sequences that comply with physical laws and ``\#long-horizon'' denotes the dataset that can serve as a long-horizon benchmark.}
\label{tab:dataset_compare}
\scalebox{0.8}{
\begin{tabular}{l|c|c|c|c|c|c}  
\toprule
 & \#Clips & \#Hours & \#Avg. Length     & \#long-horizon & \#physically-feasible & \#vision\\  
\midrule
BABEL~\citep{punnakkal2021babel}         & 52.9K & 33.2   &2.3s            & \Checkmark        & \Checkmark    & \Checkmark    \\
KIT~\citep{plappert2016kit}              & 5.7K  & 11.2   &9.5s            & \XSolidBrush      & \Checkmark    & \XSolidBrush   \\
HumanML3D~\citep{guo2022generating}      & 29.2K & 28.6   &7.1s            & \XSolidBrush      & \XSolidBrush  & \XSolidBrush   \\
Motion-X~\citep{lin2024motion}           & 81.1K & 144.2  &6.4s            & \XSolidBrush      & \XSolidBrush  & \Checkmark   \\
MotionLib~\citep{wang2024quo}            & 1.2M  & 1456.4 &-               & \XSolidBrush      & \Checkmark    & \Checkmark \\
MotionMillion~\citep{fan2025go}          & 2M    & -      &-               & \XSolidBrush      & \XSolidBrush  & \Checkmark  \\
HuMo100M~\citep{cao2025being}            & 5.7M  & 8508.3 &5.3s            & \Checkmark        & \Checkmark    & \Checkmark  \\
\midrule
OpenT2M                                  &1M     & 2815.6 &\textbf{10.1s}  & \Checkmark        & \Checkmark    & \Checkmark   \\
\bottomrule
\end{tabular}
}
\vspace{-3mm}
\end{table*}

\noindent\textbf{Multi-granularity Filtering.}
Although web videos can serve as a rich source for human motion ~\citep{kay2017kinetics,wang2023internvid,grauman2024ego}, their motion quality is often compromised by bad cases like occlusions, blur, and low resolution. 
To avoid these, we apply a set of quality criteria to ensure motions' high fidelity after extracting 2D keypoints using a pre-trained detector~\citep{xu2022vitpose}, including:
(1) A minimum keypoint count per frame to maintain structural completeness and remove occluded or partial-body sequences;
(2) A lower-bound ratio of human bounding boxes for sufficiently visible and detailed human bodies to ensure precise motion estimation and text annotation;
(3) A shortest temporal duration to exclude fragmented clips to retain continuous activities. 
These criteria help to build a high-quality motion extraction pipeline with clear video clips and semantic-rich text labels. 
Similarly, additional details about their setups can be found in Appendix~\ref{sec:additional_details_motionlib}. 

\noindent\textbf{Second-level Text Annotation.}
The quality of text annotations, in both semantic richness and precision, is critical for dataset integrity and motion generation fidelity.
Unlike prior works using a single-stage approach~\citep{cao2025being,wang2024quo} to generate a single, coarse description for an entire video clip, which fails to capture all activities within the video, our method alleviate crucial detail omission to obtain finer-grained motion alignment.
We achieve this by implementing a two-stage pipeline: Gemini-2.5-pro~\citep{team2024gemini} firstly produces temporally precise, second-by-second descriptions of human motions, including fine-grained limb movements.
These fine-grained motion descriptions are then synthesized into a coherent summary for the entire clip.
This process captures comprehensive action details, providing reliable text annotation for pretraining a robust motion generation model.

\begin{figure*}[ht]
\centering
    \includegraphics[width=1.0\linewidth]{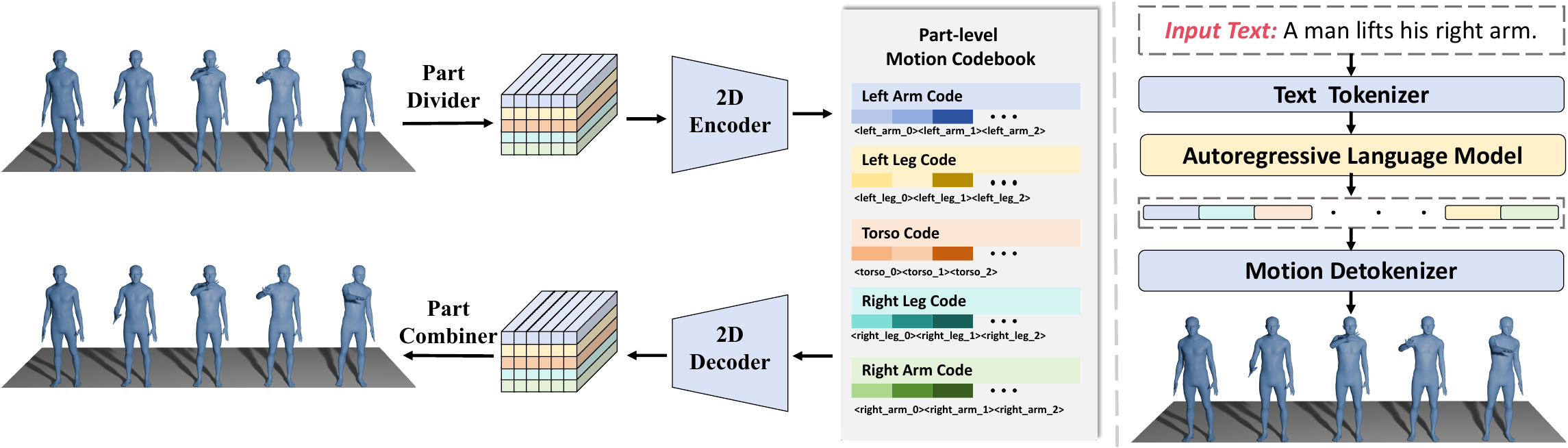}
\caption{\textbf{Model Overview.} We propose an extendable, autoregressive (AR) and discrete T2M model with no frills. 
\textbf{(left)} Our core design 2D-PRQ divides the entire body into five parts, encoding and quantizing motion into a sequence of discrete part-level tokens. 
 \textbf{(right)}
The AR model takes text as input and predicts part-level motion tokens.
We call this model ``\ModelName'' to show its simplicity.
}
\label{fig: framework}
\vspace{-3mm}
\end{figure*}

\noindent\textbf{Long-horizon Motion Curation.}
Most existing motion data is predominantly short-duration, limiting their utility for long-horizon benchmarks.
While works like BABEL~\citep{punnakkal2021babel} have explored to address this issue, their scale and duration still remain constrained. 
Here, we curate a strategy to synthesize massive long-horizon motion sequences.
First, we concatenate raw motions via interpolation with orientation and global coordinate alignment. 
However, such intuitive operation can lead to physically implausible transitions.
Therefore, we replace it with a two-step refinement: an RL-based policy filters out untrackable motions, and the avatars' trajectories are used to ensure physically feasible transitions. 
In addition, previous works~\citep{cao2025being} create long-horizon text by directly concatenation as well, which introduces noise and inefficiency due to motion-irrelevant content. 
Instead, we use Gemini-2.5-pro to merge refined concise commands (e.g., “wave left hand.”) into clean, user-friendly descriptions.
As far as we know, \DataName is the first dataset with an average motion length exceeding 10 seconds.
A statistical comparison with counterparts is illustrated in Table~\ref{tab:dataset_compare}.

\section{The MonoFrill Model}
\noindent\textbf{Overview.}
Inspired by large language models' success in multimodal understanding~\citep{luo2020univl,zhang2024pixels,touvron2023llama,zhang2024video}, we frame human motion as a specialized ``language''.
Our approach, illustrated in Figure~\ref{fig: framework}, uses a motion tokenizer to discretize sequences into tokens, which are then generated autoregressively by an LLM.
To integrate motion tokens into the LLM backbone, we expand the LLM's vocabulary by incorporating the $K$ discrete codes. 
We also introduce special tokens such as $<$mot$>$, $<$/mot$>$ to  delimit motion sequences.
The overall training pipeline consists of two phases. First, we train a motion tokenizer to discretize motion features into motion tokens while minimizing reconstruction error. This is followed by a text-motion alignment training via motion instruction tuning~\citep{jiang2023motiongpt}, which is conducted on \DataName to achieve robust and general-purpose text-motion alignment. 
We name our model as ``\ModelName'' to denote its simplicity and extendable capability without any complex design. 

\noindent\textbf{Motion Instruction Training.}
Achieving robust text-motion alignment is essential for developing a generalizable motion generation model. 
In the text-alignment training phase, 2D-PRQ first encodes and quantizes the continuous raw motion features $\mathcal{M} \in \mathbbm{R}^{T\times D}$ sequence into discrete motion tokens $\mathcal{V} \in \mathbbm{R}^{n \times p \times l}$, using a temporal downsampling ratio of $n/T$. 
Here, $p=5$ represents the number of body parts, $n$ is the number of temporal tokens, and $l$ is the number of residual layers in the quantization process, $K$ is the size of the motion codebook.
In addition to common motion tokens, we also introduce another two special tokens $<$part$>$, and $<$/part$>$ to separate body-part-specific subsequences in order to structure the input effectively.
To enable autoregressive prediction of motion tokens conditioned on descriptions, we design a standardized template for all text-motion pairs:
\begin{tcolorbox}[mybox]
\vspace{-2mm}
\textbf{Input $\mathcal{I}$}: The person performs a salute and then shakes hands with another person. \\
\textbf{Answer $\mathcal{M}$}: $<$mot$>$ $<$part\_{1}$>$$<$motion\_token$>$ ... $</$part\_{1}$>$ ... $</$mot$>$
\vspace{-2mm}
\end{tcolorbox}
To train our large motion model, we optimize the negative log-likelihood loss over the predicted tokens as follows:
\begin{equation}
    \mathcal{L}(\Theta)=-\sum_{j=1}^{L} \log P_{\Theta}(y_j|desc, \hat{y}_{1:j-1}),
\end{equation}
where $\hat{y}$ and $y$ denote the input and target token sequences, respectively. $\Theta$ represents the model parameters and $L$ is the length of the target sequence. $desc$ represents text input.

\noindent\textbf{2D-PRQ: Towards Generalized Motion Tokenization.}
The increasing scale of motion datasets demands more effective encoding.
Current VQ-based methods~\citep{zhang2023motiongpt,zhang2023generating} use 1D temporal convolutions and a single embedding for the whole body, leading to information loss and limited generalization.
In this work, we propose 2D-PRQ, a novel tokenizer that captures spatiotemporal dependencies by decomposing the body into parts.
Given a motion sequence $m_{1:T} \in \mathbbm{R}^{T\times D}$, 2D-PRQ first splits it into part-level features $\tilde{m}_{1:T} \in \mathbbm{R}^{T\times p\times d}$, where $d$ is the part-level feature dimension, and $p$=$5$ represents the body parts: $\{$left arm, left leg, torso, right leg, right arm$\}$.
More details about how to split the whole body feature into independent parts can be found in Appendix \ref{sec:split_part}. Unlike methods that process parts in isolation~\citep{chen2025language}, we conceptualize the sequence as a 2D image: time as width and body parts as height.
Such design allows us to use a 2D convolution block for motion encoding\cite{he2016deep}, capturing both temporal correlations across frames and spatial dependency between different body parts, which is crucial for maintaining whole-body coordination and consistency.
The encoder outputs a latent sequence $\tilde{b}_{1:p;1:n}$ with a downsampling ratio of $n/T$.
Each latent vector $\tilde{b}_{i,j}$ is quantized via residual quantization~\citep{lee2022autoregressive} using a shared codebook $\mathbbm{C}$, producing the token sequence $[b_{1:p;1:n}^k]^K_{k=0}$, where $b^k$ denotes the code sequence at layer $k$.
For the decoding, a symmetric 2D decoder reconstructs the part-level features $\hat{m}_{1:p;1:n}$ which are aggregated to restore the raw motion feature $\hat{m}_j$.
The reconstruction loss is:
\begin{equation}
    \mathcal{L} = ||m-\hat{m}||_1+\sum\limits_{i=0}^p||m_i-\hat{m}_i||_1+\beta\sum\limits_{k=1}^K\sum\limits_{i=1}^p||r^k_i-sg[b^k_i]||_2^2.
\end{equation}
\section{Experiments}
\subsection{Experimental Setup}

\textbf{Datasets.} 
To evaluate the performance and generalization capabilities of our model, we conduct experiments on three diverse motion datasets: HumanML3D~\citep{guo2022generating}, Motion-X~\citep{lin2024motion}, and our collected \DataName.
HumanML3D is a widely adopted benchmark for text-to-motion generation, comprising 4,616 high-quality motion sequences paired with 44,970 textual annotations derived from sources like the AMASS dataset.
Motion-X extends this scale with approximately 81,000 motion sequences, incorporating multi-modal data (e.g., video and audio cues) to enhance diversity in complex interactions and long-horizon motions. 
For further validation on an even larger scale, we utilize \DataName, a comprehensive dataset with over 1 million motion sequences sourced from real-world human activities, which covers a broad spectrum of human activities, such as walking, dancing, and sports, making it ideal for assessing motion synthesis from diverse language descriptions. 
Following established protocols, we partition each dataset into training, validation, and test splits using an 80\%, 5\%, and 15\% ratio, respectively.

\noindent\textbf{Evaluation Metrics.}
Our experiments center on two primary tasks to comprehensively assess \ModelName on text-to-motion (T2M) generation and 2D-PRQ on motion reconstruction.
For T2M generation, we adopt standard metrics from the literature~\citep{guo2022generating}, including Motion-retrieval Precision (R-Precision), Multimodal Distance (MMDist), and Frechet Inception Distance (FID).
In addition, the effect of motion tokenizers is assessed by the motion reconstruction task, which reconstructs input motions through the tokenizer to verify discretization quality.
We employ FID to measure overall sequence realism and Mean Per Joint Position Error (MPJPE) to quantify geometric accuracy.
Details of these metrics can be seen in Appendix~\ref{sec:Evaluation_Metrics}.

\begin{table*}[h]
\centering
\renewcommand{\arraystretch}{1.1}
\setlength{\tabcolsep}{12pt}
\caption{Comparison of zero-shot performance on \DataName$_{\rm zero}$ using different datasets for training. Models trained on \DataName consistently present significant OOD improvements.}
\label{tab:ood}
\vspace{-2mm}
\scalebox{0.75}{
\begin{tabular}{cc|cccccc}
\midrule
\#Model & \#training data & R@1 $\uparrow$ & R@2 $\uparrow$ & R@3 $\uparrow$ & FID $\downarrow$ & MMDist $\downarrow$ & DIV $\uparrow$ \\
\midrule
Real & -  & 0.316 & 0.495 & 0.621 & - & 3.771 & 7.749  \\
\midrule
\rowcolor{BlockA!30}
$\text{MDM}$                      &HumanML3D    &0.065          &0.126          &0.180          &51.307         &7.642            &3.040   \\
\rowcolor{BlockA!30}
$\text{MDM}$                      &Motion-X     &0.055          &0.107          &0.160          &56.257         &8.008            &3.019   \\
\rowcolor{BlockA!30}
$\text{MDM}$                      &OpenT2M      &\textbf{0.194} &\textbf{0.338} &\textbf{0.447} &\textbf{8.153}  &\textbf{4.889}  &\textbf{7.136}   \\
\midrule
\rowcolor{BlockA!30}
$\text{T2M-GPT}$                  &HumanML3D    &0.070          &0.130          &0.186           &62.036         &8.093           &2.586   \\
\rowcolor{BlockA!30}
$\text{T2M-GPT}$                  &Motion-X     &0.063          &0.120          &0.173           &53.464         &7.770           &2.957   \\
\rowcolor{BlockA!30} 
$\text{T2M-GPT}$                  &OpenT2M      &\textbf{0.159} &\textbf{0.271} &\textbf{0.357}  &\textbf{5.566} &\textbf{5.072}  &\textbf{6.921} \\
\midrule
\rowcolor{BlockA!30}
$\text{Being-M0}$                 &HumanML3D    &0.073           &0.134          &0.190           &58.541         &7.956           &2.932   \\
\rowcolor{BlockA!30}
$\text{Being-M0}$                 &Motion-X     &0.057           &0.109          &0.157           &46.222         &7.652           &3.220   \\
\rowcolor{BlockA!30}
$\text{Being-M0}$                 &OpenT2M      &\textbf{0.155}  &\textbf{0.266} &\textbf{0.356}  &\textbf{5.811} &\textbf{5.110}  &\textbf{7.090}   \\
\midrule
\rowcolor{BlockB!30}
$\text{\ModelName-2D-PRQ}_{4}$    &HumanML3D    &0.061           &0.119          &0.173           &60.177         &8.059           &2.674   \\
\rowcolor{BlockB!30}
$\text{\ModelName-2D-PRQ}_{4}$    &Motion-X     &0.052           &0.110          &0.152           &55.470         &7.841           &2.433   \\
\rowcolor{BlockB!30}
$\text{\ModelName-2D-PRQ}_{4}$    &OpenT2M      &\textbf{0.240}  &\textbf{0.399} &\textbf{0.512}  &\textbf{1.475} &\textbf{4.281}  &\textbf{7.563}   \\
\bottomrule
\end{tabular}}
\end{table*}

\begin{table*}[!ht]
\centering
\renewcommand{\arraystretch}{1.1}
\setlength{\tabcolsep}{12pt}
\caption{Comparison of motion instruction tuning on HumanML3D. 
We apply a limited number of training steps to avoid overfitting.
Models with $\text{\#pretrain}$ consistently achieve significant improvements across diverse \text{\#LLM backbones}.}
\label{tab:pretrain_dataset}
\vspace{-2mm}
\scalebox{0.75}{
\begin{tabular}{ccc|cccccc}
\midrule
\#Model & \#LLM backbone & \#pretrain & R@1 $\uparrow$ & R@2 $\uparrow$ & R@3 $\uparrow$ & FID $\downarrow$ & MMDist $\downarrow$ & DIV $\uparrow$ \\
\midrule
Real & - & - & 0.519 & 0.710 & 0.801 & - & 3.176 & 10.954  \\
\midrule
\rowcolor{BlockA!30}
\ModelName    &GPT2-medium  &-           &0.078  &0.148   &0.212  &61.809  &8.803 & 4.810     \\
\rowcolor{BlockA!30}
\ModelName    &LlaMA2-7B    &-           &0.472  &0.645   &0.741  &0.619   &3.572 &11.226     \\
\rowcolor{BlockA!30}
\ModelName    &LlaMA3-8B    &-           &0.503  &0.694   &0.792  &0.546   &3.224  &11.104   \\
\midrule
\rowcolor{BlockB!30}
\ModelName    &GPT2-medium  &\checkmark  &0.215  &0.316           &0.377           &17.91           &7.129           &8.372      \\
\rowcolor{BlockB!30}
\ModelName    &LlaMA2-7B    &\checkmark  &0.485  &0.676           &0.773           &0.435           &3.386           &11.373    \\
\rowcolor{BlockB!30}
\ModelName    &LlaMA3-8B    &\checkmark  &\textbf{0.518}   &\textbf{0.704}  &\textbf{0.798}  &\textbf{0.238}  &\textbf{3.172}  &\textbf{11.216}   \\
\bottomrule

\end{tabular}}
\vspace{-3mm}
\end{table*}

\noindent\textbf{Implementation Details.}
For the motion reconstruction task, we implement a motion encoder with a temporal downsampling rate of $\alpha = 4$ for fair comparison. The motion tokenizer is trained with a learning rate of 2e-4 and a batch size of 256. 
We implement our $\text{\ModelName-2D-PRQ}_{4}$ with three sizes of LLMs: GPT2-medium~\citep{lagler2013gpt2}, LlaMA2-7B~\citep{touvron2023llama2}, and LlaMA3.1-8B~\citep{dubey2024llama}. 
Full parameter training is performed on 8 $\times$ A800 GPU with a learning rate of 2e-4 and a batch size of 1024 over 5000 steps on \DataName.

\subsection{Effectiveness of OpenT2M Dataset}

While previous works have introduced large-scale datasets \citep{fan2025go, wang2024quo,xu2024motionbank}, their impact on model capabilities still remains inadequately explored. To valid the effectiveness of \DataName, we conduct a rigorous T2M evaluation focusing on the following key aspects: (1) zero-shot generalization to out-of-domain cases, (2) adaptation to novel motion activities via instruction tuning, and (3) long-horizon motion generation.

\noindent\textbf{Zero-shot Motion Generalization.} 
To rigorously assess the generalization of T2M models to unseen data, we curate a held-out evaluation set \DataName$_{\rm zero}$ comprising 12,000 motions excluded from training data, including HumanML3D and \DataName, ensuring no domain overlap between the evaluation and training sets. 
This OOD benchmark enables zero-shot evaluation, where models generate motions for novel text prompts without task-specific fine-tuning.
We benchmark three representative baselines: MDM~\citep{tevet2022human}, T2M-GPT~\citep{zhang2023generating}, Being-M0~\citep{wang2024quo}, as well as our \ModelName.
As shown in Table \ref{tab:ood}, models trained on HumanML3D and Motion-X exhibit limited zero-shot performance, with metrics like FID and R-Precision revealing degraded semantic alignment and motion diversity OOD sequences.
In contrast, training on \DataName yields substantial improvements across all baselines, underscoring its role in enhancing generalization through diverse, large-scale coverage of motion primitives and contexts.

\noindent\textbf{Motion Instruction Tuning.}
Inspired by the two-stage training paradigm in multimodal vision-language models~\citep{liu2023improvedllava}, we adopt a similar pipeline for T2M generation: an initial pretraining phase on our large-scale \DataName dataset to foster robust text-motion alignment, followed by targeted fine-tuning on downstream benchmarks.
Specifically, we fine-tune the pre-trained model on HumanML3D for a limited 50 epochs.
Unlike previous works that train for up to 300 epochs on the same dataset --- potentially leading to in-domain overfitting --- we intentionally restrict the number of training steps. 
This allows us to assess inherent generalization capabilities without conflating them with the effects of prolonged training, a potential confound in prior evaluations.
As shown in Table~\ref{tab:pretrain_dataset}, models pre-trained on \DataName consistently outperform their non-pre-trained counterparts, indicating that pre-training equips the model with generalized motion patterns.

\begin{table*}[!ht]
\centering
\renewcommand{\arraystretch}{1.1}
\setlength{\tabcolsep}{12pt}
\caption{\textbf{Comparison on \DataName$_{\rm long}$}, where ``\#text refinement'' refers to converting raw texts into cleaned user commands, "\#long-horizon" denotes incorporating long-horizon motion data into \DataName.}
\label{tab:benchmark_long_term}
\vspace{-2mm}
\scalebox{0.75}{
\begin{tabular}{ccc|cccccc}
\midrule
Model & \#text refinement & \#long-horizon & R@1 $\uparrow$ & R@2 $\uparrow$ & R@3 $\uparrow$ & FID $\downarrow$ & MMDist $\downarrow$ & DIV $\uparrow$ \\
\midrule
Real          &\checkmark   &\checkmark           &0.573   &0.740        &0.822             &-                &2.842             &10.450  \\
\midrule
\rowcolor{BlockA!30}
\ModelName    &-            &-           &0.091   &0.165        &0.226             &36.837           &7.976             &5.871           \\
\rowcolor{BlockA!30}
\ModelName    &-            &\checkmark  &0.484  &0.648         &0.738             &0.430            &3.520             &10.682           \\
\rowcolor{BlockB!30}
\midrule
\ModelName    &\checkmark   &\checkmark  &\textbf{0.510} &\textbf{0.677} &\textbf{0.765}    &\textbf{0.297}   &\textbf{3.322}    &\textbf{10.748}   \\
\bottomrule
\end{tabular}}
\end{table*}

\begin{figure*}[ht]
\centering
    \includegraphics[width=0.9\linewidth]{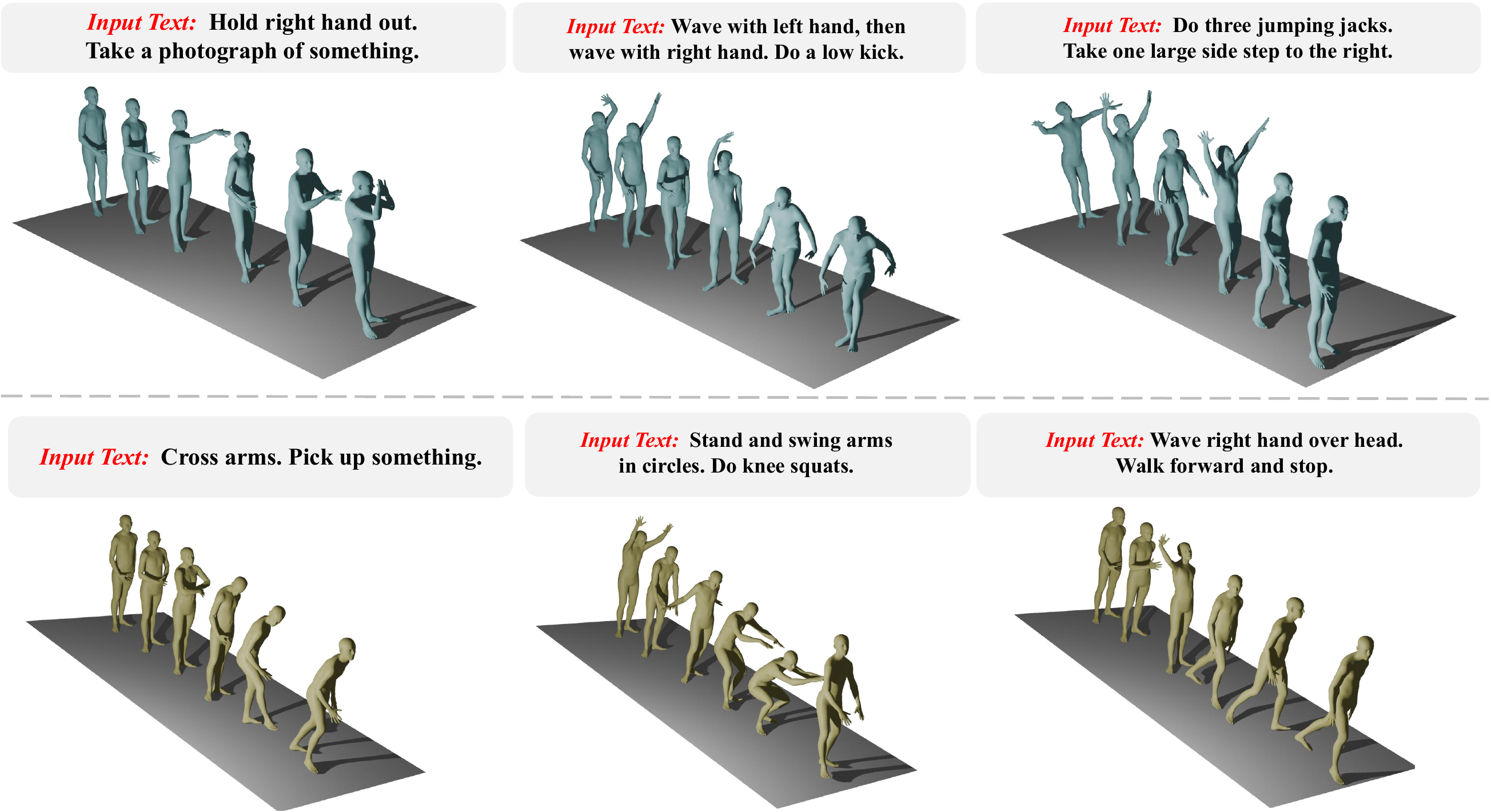}
 \caption{\textbf{Visualization of generated long-horizon motions.} Visualization results demonstrate the ability to generate long-horizon motion sequences that accurately align with complex texts.}
\label{fig:long_vis}
\vspace{-5mm}
\end{figure*}

\noindent\textbf{Long-horizon Motion Generation.}
Before introducing long-horizon benchmark, we first conduct text refinement. Text annotations in existing datasets, such as HumanML3D, contain considerable redundant details. Directly concatenating texts to construct long-horizon benchmark will introduce noise and inefficiency due to motion-irrelevant content. To mitigate this issue, we design a specific prompt and utilize Gemini-2.5 ~\citep{team2024gemini} to conduct text refinement: (1) removing motion-irrelevant details; (2) converting text annotations into cleaned and precise user commands. As illustrated in Table \ref{tab:text_refine}, this text refinement results in an improvement in R-Precision, achieving a better alignment between the refined text and motion sequences. 

\begin{wraptable}{r}{8cm}
\vspace{-6mm} 
\centering
\renewcommand{\arraystretch}{1.1}
\setlength{\tabcolsep}{12pt}
\caption{Ablation of text refinement on HumanML3D}
\label{tab:text_refine}
\scalebox{0.75}{
\begin{tabular}{c|ccc}
\toprule
\#text refinement & R@1 $\uparrow$   & R@2 $\uparrow$ & R@3 $\uparrow$ \\
\midrule
-     &0.520             &0.709            &0.801         \\
\checkmark  &\textbf{0.533}    &\textbf{0.720}   &\textbf{0.808}     \\
\bottomrule
\end{tabular}}
\vspace{-2mm}
\end{wraptable}

Following text refinement, we introduce \DataName$_{\rm long}$, a long-horizon benchmark built with our curation pipeline to evaluate T2M models on extended sequence generation. 
Our evaluation of a leading model, \ModelName, reveals a significant struggle to produce satisfied performance without training on long-horizon motion data.
In addition, text refinement further substantially improves this ability by enhancing text-motion alignment.
Visualizations of the generated sequences are provided in Figure \ref{fig:long_vis},  and a detailed comparison with the BABEL dataset is available in Appendix \ref{sec:Long_horizon_motion_Comparison}.

\begin{table*}[!ht]
\centering
\renewcommand{\arraystretch}{1.1}
\setlength{\tabcolsep}{12pt}
\caption{Comparison of motion reconstruction on three benchmarks. Subscripts denote the number of quantization layers.}
\label{tab:vq_vae}
\vspace{-2mm}
\scalebox{0.75}{
\begin{tabular}{c|c|cc|cc|cc}
\toprule
\multicolumn{2}{c}{}&\multicolumn{2}{|c}{HumanML3D} &\multicolumn{2}{|c}{Motion-X} &\multicolumn{2}{|c}{OpenT2M}\\
\midrule
Motion Tokenizer &Codebook Size &FID $\downarrow$ & MPJPE $\downarrow$ &FID $\downarrow$ & MPJPE $\downarrow$  &FID $\downarrow$ & MPJPE $\downarrow$ \\
\midrule
\rowcolor{BlockA!30}
$\text{VQ-VAE}_{1}$  &1024   &0.358            &83.902          &0.127           &115.382         &3.130           &178.534         \\
\rowcolor{BlockA!30}
$\text{FSQ}_{1}$     &65536  &0.151            &70.480          &0.828           &110.021         &1.962           &165.084         \\
\rowcolor{BlockA!30}
$\text{RQ-VAE}_{6}$  &1024   &0.031            &48.696          &0.013           &67.390          &0.080             &96.753              \\
\rowcolor{BlockA!30}
$\text{RQ-VAE}_{8}$  &1024   &0.021            &45.633          &0.020           &65.484          &0.062            &84.655              \\
\rowcolor{BlockA!30}
$\text{PRQ}_{4}$     &1024   &0.003            &28.703          &0.012           &73.989          &0.094            &95.743              \\
\rowcolor{BlockA!30}
$\text{PRQ}_{6}$     &1024   &0.005            &25.485          &0.009           &58.155          &0.029            &67.569             \\
\midrule
\rowcolor{BlockB!30}
$\text{2D-PRQ}_{4}$ &1024   &\textbf{0.003}   &28.628           &0.011           &54.493          &0.022            &49.134          \\
\rowcolor{BlockB!30}
$\text{2D-PRQ}_{6}$ &1024   &0.005            &\textbf{25.417}  &\textbf{0.008}  &\textbf{48.099} &\textbf{0.021}   &\textbf{37.922}             \\
\bottomrule
\end{tabular}}
\end{table*}
\begin{table*}[h]
\centering
\renewcommand{\arraystretch}{1.1}
\setlength{\tabcolsep}{12pt}
\caption{Comparison of T2M on \DataName under different model parameters and motion tokenizers.}
\label{tab:benchmark_motiondb_all}
\vspace{-2mm}
\scalebox{0.75}{
\begin{tabular}{c|c|cccccc}
\midrule
Model & LLM & R@1 $\uparrow$ & R@2 $\uparrow$ & R@3 $\uparrow$ & FID $\downarrow$ & MMDist $\downarrow$ & DIV $\uparrow$ \\
\midrule
\rowcolor{BlockA!30}
$\text{\ModelName-VQ}_{1}$      &GPT2-medium  &0.257          &0.410           &0.513           &11.226          &5.146           &7.393   \\
\rowcolor{BlockA!30}
$\text{\ModelName-VQ}_{1}$      &LlaMA2-7B    &0.345          &0.534           &0.656           &3.005           &3.955           &8.463   \\
\rowcolor{BlockA!30}
$\text{\ModelName-VQ}_{1}$      &LlaMA3-8B    &0.345          &0.534           &0.656           &2.979           &3.960           &8.437   \\
\midrule
\rowcolor{BlockB!30}
$\text{\ModelName-2D-PRQ}_{4}$  &GPT2-medium  &0.357          &0.534           &0.645           &8.880           &4.316           &7.905       \\
\rowcolor{BlockB!30}
$\text{\ModelName-2D-PRQ}_{4}$  &LlaMA2-7B    &\textbf{0.491} &\textbf{0.675}  &\textbf{0.777}  &\textbf{0.475}  &\textbf{2.962}  &\textbf{9.450}   \\
\rowcolor{BlockB!30}
$\text{\ModelName-2D-PRQ}_{4}$  &LlaMA3-8B    &0.478          &0.668           &0.777           &0.552           &3.012           &8.901   \\
\bottomrule
\end{tabular}}
\vspace{-3mm}
\end{table*}

\subsection{Effectiveness of 2D-PRQ}
\textbf{Comparison of Motion Reconstruction.}
As shown in Table \ref{tab:vq_vae}, 2D-PRQ outperforms previous methods, including PRQ, on large-scale datasets.
Under a consistent configuration (codebook size 1024, feature dim 512, except for FSQ~\citep{mentzer2023finite}) (codebook size 65536), 2D-PRQ achieves substantially lower reconstruction error on Motion-X and \DataName while using a simpler architecture.
The key advantage lies in its 2D convolutional design, which jointly models spatial and temporal dependencies. 
This leads to marginal gains on HumanML3D (MPJPE: 25.417 vs. 25.485) but dramatically larger improvements as the dataset scale increases, as evidenced by results on Motion-X (MPJPE: 54.493 vs. 73.989) and \DataName (MPJPE: 49.134 vs. 95.743).

\noindent\textbf{Comparison of Text-to-Motion Generation.}
The choice of motion tokenizer is critically dependent on the scale of the training data.
As shown in Table~\ref{tab:ood}, replacing VQ-VAE with our 2D-PRQ tokenizer in the Being-M0 model leads to a performance drop when training on smaller datasets like HumanML3D and Motion-X.
We attribute this to the increased number of motion tokens in 2D-PRQ, which requires large-scale data for effective training.
This hypothesis is confirmed when training on the large-scale \DataName: here, the $\text{\ModelName-2D-PRQ}_{4}$ model achieves superior zero-shot performance, even exceeding strong baselines like T2M-GPT, Being-M0, and MDM.
This result, also evident in Table \ref{tab:benchmark_motiondb_all}, underscores that 2D-PRQ unlocks the full potential of large datasets and highlights the critical role of a well-designed motion representation.
In Table~\ref{tab:benchmark_motiondb_all}, we observe that scaling the LLM from GPT2-medium to LLaMA2-7B brings significant gains. However, further scaling to LLaMA3-8B yields diminishing returns. This phenomenon can be found both VQ-VAE and 2D-PRQ, indicating that the performance degradation is unrelated to the interaction between the backbone and the tokenizer. Furthermore, we perform hyperparameter tuning and observe that scaling up the backbone from Llama2-7B to Llama3-8B still does not yield significant gains. Therefore, we hypothesize a saturation point where performance becomes less dependent on LLM size. 

\begin{wraptable}{r}{8cm}
\vspace{-2mm} 
\centering
\renewcommand{\arraystretch}{1.1}
\caption{Zero-shot comparison of motion tokenizers.}
\label{tab:vq_vae_generalization}
\scalebox{0.75}{
\begin{tabular}{c|cc|cc}
  \toprule
  \multicolumn{1}{c}{} & \multicolumn{2}{|c}{HumanML3D} & \multicolumn{2}{|c}{Motion-X} \\
  \midrule
  Motion Tokenizer     & FID $\downarrow$ & MPJPE $\downarrow$ & FID $\downarrow$ & MPJPE $\downarrow$ \\
  \midrule
  \rowcolor{BlockA!30}
  $\text{VQ-VAE}_{1}$    &25.525           &237.702             &44.889           &293.301 \\
  \rowcolor{BlockA!30}
  $\text{PRQ}_{4}$       &2.169            &135.964              &5.020           &167.508 \\
  \midrule
  \rowcolor{BlockB!30}
  $\text{2D-PRQ}_{4}$    &\textbf{0.107}   &\textbf{77.695}      &\textbf{1.606}  &\textbf{108.921} \\
  \bottomrule
\end{tabular}}
\vspace{-2mm}
\end{wraptable}

\noindent\textbf{Comparison under Zero-shot Setup.} 
Previous work primarily adopts the $\text{VQ-VAE}_{1}$ tokenizer and trains it on limited-scale datasets for extensive periods (e.g., 200K steps), which leads to overfitting and fails to assess the tokenizer's inherent zero-shot performance.
In contrast, we pre-train various tokenizers on the large-scale \DataName dataset and evaluate their zero-shot performance on HumanML3D and Motion-X.
As shown in Table \ref{tab:vq_vae_generalization}, $\text{2D-PRQ}_{4}$ significantly shows superior zero-shot performance compared with $\text{VQ-VAE}_{1}$. Furthermore, compared with $\text{PRQ}_{4}$, $\text{2D-PRQ}_{4}$ reduce reconstruction error from 135.964 to 77.695 per frame on HumanML3D, demonstrating the superior generalization and effectiveness in mitigating tokenizer overfitting.

\section{Conclusion}
This paper introduces \textbf{\DataName}, a large-scale, high-quality human motion dataset with physically feasible validation, multi-granularity filtering, and second-wise annotation. We also introduce a pipeline that synthesizes long-horizon motion autonomously, containing motion connection and text connection to equip T2M models with the capability to generate complex and long-horizon motion sequences. Leveraging \DataName, we introduce \ModelName, a pretrained T2M model achieving superior performance without complicated  ``frills''. As the core component of \ModelName, \textbf{2D-PRQ}, a novel motion tokenizer, decouples human body features into five parts and captures spatiotemporal dependencies by applying 2D convolution, showing superior reconstruction performance on large-scale datasets and zero-shot ability. Comprehensive experiments demonstrate that \DataName shows benefits in improving generalization on unseen motion sequences and motion instruction tuning. We hope that our findings and the release of \DataName will benefit this field.

\clearpage

\bibliographystyle{plainnat}
\bibliography{ref}

\clearpage

\beginappendix

In this supplementary, we provide additional details of \DataName in Section \ref{sec:additional_details_motionlib}. We provide the details about dividing human body into independent parts in Section ~\ref {sec:split_part}. We also provide details of evaluation metrics in Section ~\ref{sec:Evaluation_Metrics}. We provide visualization examples of \DataName in Section \ref{sec:visualiztion_examples}.

\section{Additional Analysis of OpenT2M}
\label{sec:additional_details_motionlib}

\subsection{Data Distribution}
\label{sec:Statistical_Analysis}
\begin{figure*}[ht]
    \centering
    \begin{subfigure}{0.48\textwidth} 
        \centering
        \includegraphics[width=\linewidth]{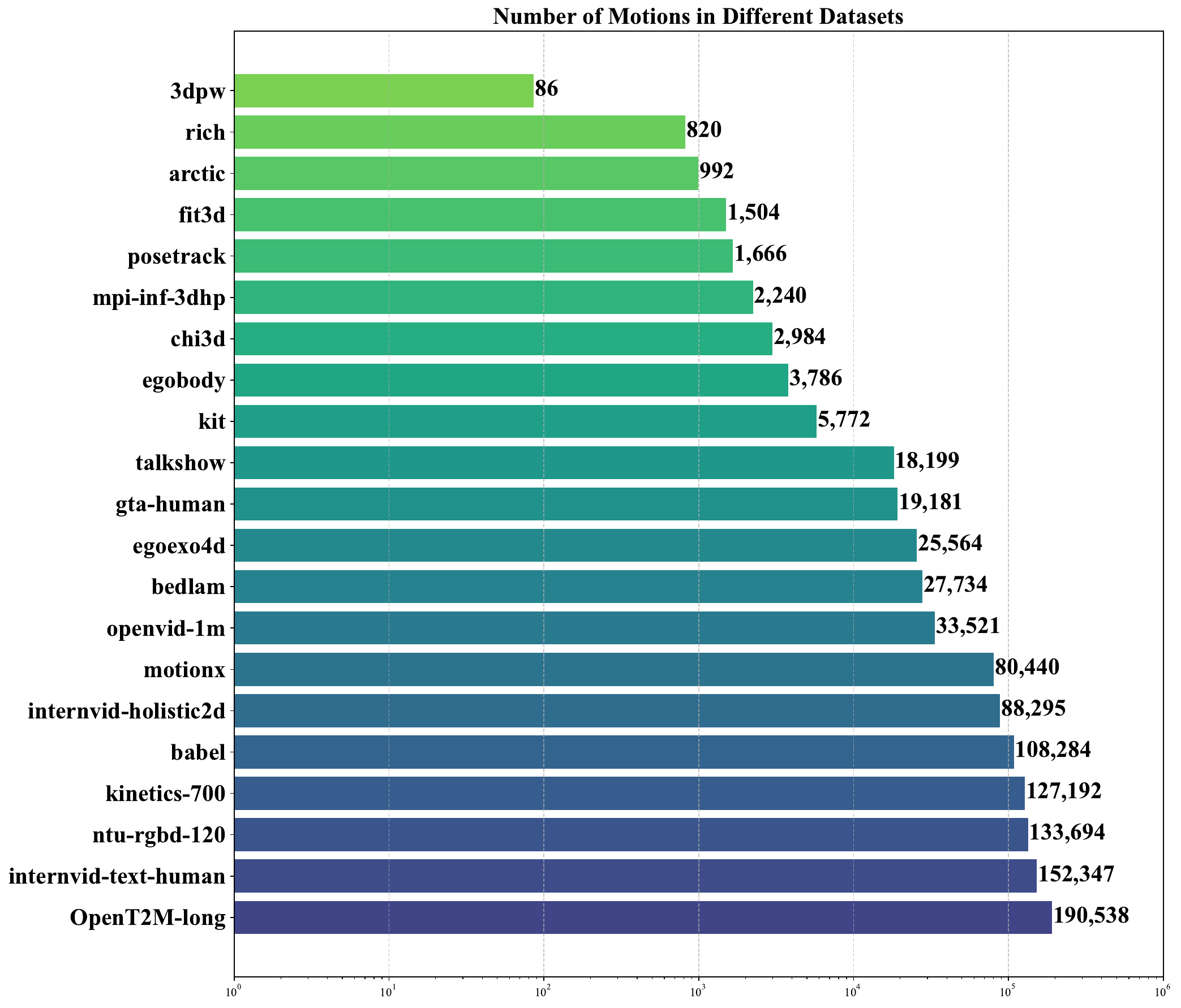}
        \caption{Distribution of motion sequences across different subsets in \DataName.}
        \label{fig:app_dataset_motions_statistics}
    \end{subfigure}
    \hfill
    \begin{subfigure}{0.48\textwidth}
        \centering
        \includegraphics[width=\linewidth]{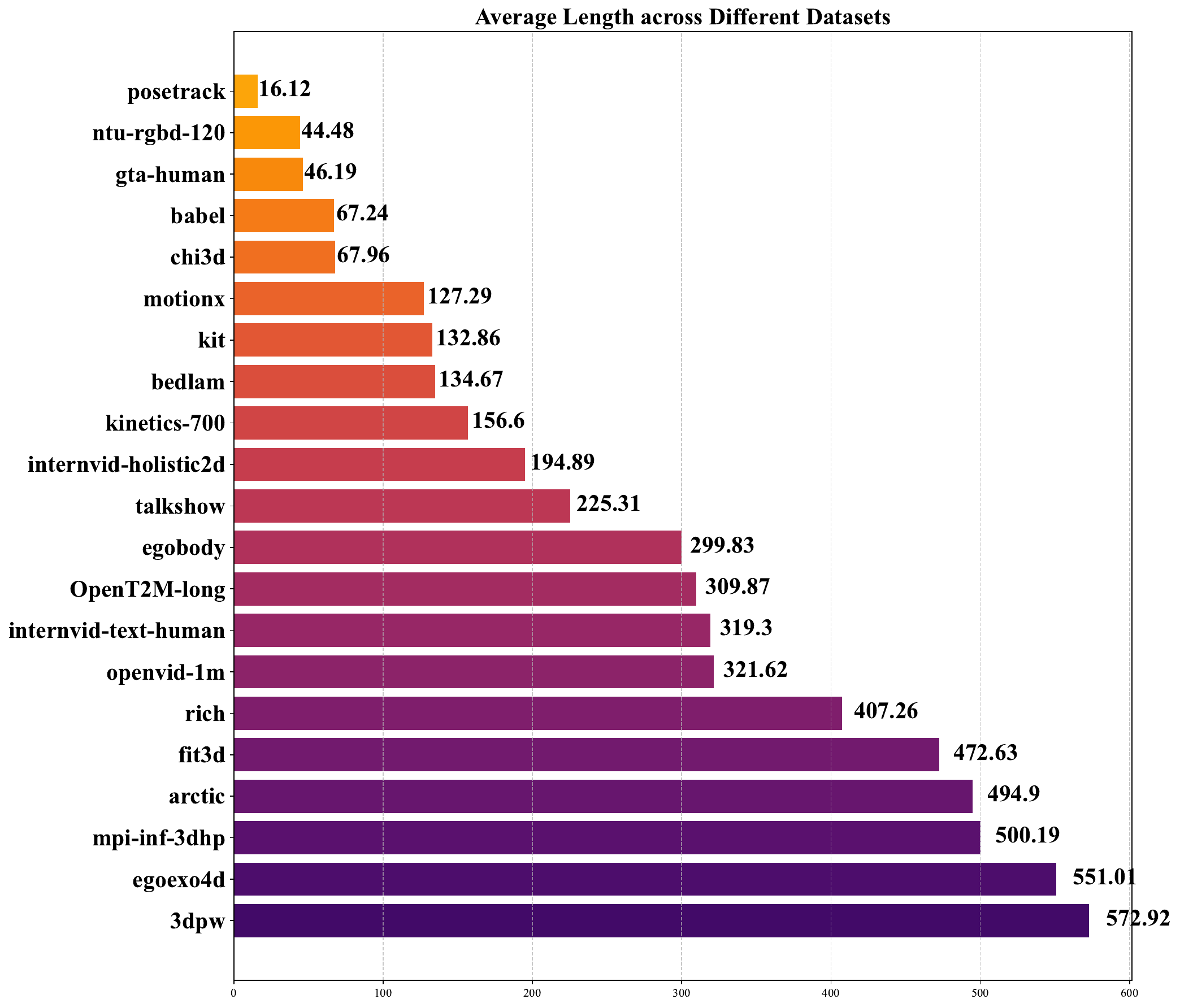}
        \caption{Average length distribution of \DataName across different subsets.}
        \label{fig:app_dataset_average_length}
    \end{subfigure}
    \caption{Statistics of the \DataName dataset. (a) Motion sequence distribution (log scale). (b) Average motion length distribution.}
    \label{fig:dataset_statistics}
\end{figure*}

Figure \ref{fig:app_dataset_motions_statistics} shows the number distribution of motion sequences across different subsets in \DataName on a logarithmic scale, demonstrating variations in dataset sizes. \DataName integrates 21 curated subsets, amounting to a comprehensive collection of 1 million motion sequences. A substantial portion of motions in \DataName are extracted from web videos utilizing motion estimation models \citep{shin2024wham}, such as Kinetics-700 \citep{kay2017kinetics}, Internvid \citep{wang2023internvid}. These motions undergo rigorous physically feasible validation and multi-granularity filtering. We set the number of visible keypoints to 8, while the whole body corresponds to 17 keypoints. Each motion sequence accounts for over 50\% of the duration of the corresponding original video, ensuring temporal consistency and semantic validity. \DataName also integrates open-source human motion datasets \citep{andriluka2018posetrack, fieraru2021aifit,cai2024playing}, such as Motion-X~\citep{lin2024motion}. Leveraging the proposed long-horizon motion curation pipeline, we construct 190K long-horizon motion sequences. The \DataName$_{\rm long}$ comprises motions spliced from two, three, four, and five individual motion sequences. Figure \ref{fig:app_dataset_average_length} shows the average length distribution of \DataName across different subsets. We observe that the subset with the shortest average sequence length is Postrack, comprising merely 16.12 frames, while 3DPW exhibits the longest average length, exceeding 500 frames. Following a meticulous curation process, \DataName exhibits a substantially longer average length compared with previous work~\citep{cao2025being}.

\subsection{Comparison of Long-horizon Datasets}
\label{sec:Long_horizon_motion_Comparison}

We first detail the pipeline for long-horizon motion curation. Two different motion sequences are initially aligned in orientation by rotating the initial frame of the second sequence to match the facing direction of the last frame in the first sequence. Subsequently, the entire second sequence is translated spatially to align its position with that of the last frame of the first sequence. Finally, a fixed transition duration is applied, during which spherical linear interpolation is performed between the last frame of the first motion and the initial frame of the second motion to ensure smooth kinematic continuity. To ensure that long-horizon motion sequences adhere to physical constraints, we utilize the concatenated motion sequence as reference poses for an RL policy, driving the avatar in the IsaacGym to track the reference motion. The resulting motion, refined through physical simulation, is adopted as the final long-horizon motion sequences.

\begin{figure*}[htbp]
    \centering
    \begin{subfigure}{0.4\textwidth}
        \centering
        \includegraphics[width=\linewidth]{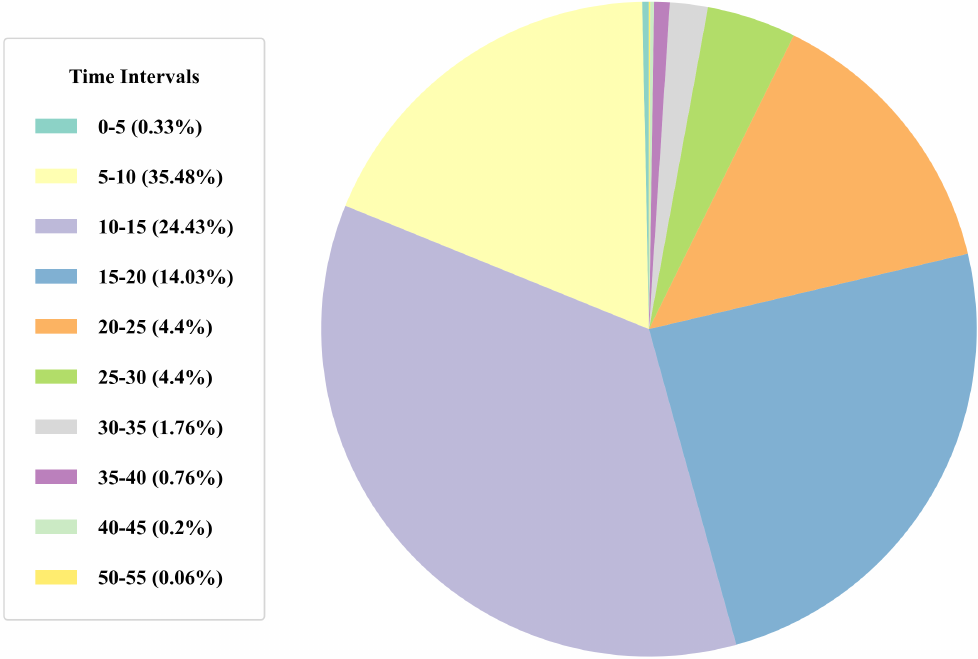}
        \caption{\DataName$_{\rm long}$ Length Distribution}
        \label{fig:humanml}
    \end{subfigure}
    \hfill
    \begin{subfigure}{0.4\textwidth}
        \centering
        \includegraphics[width=\linewidth]{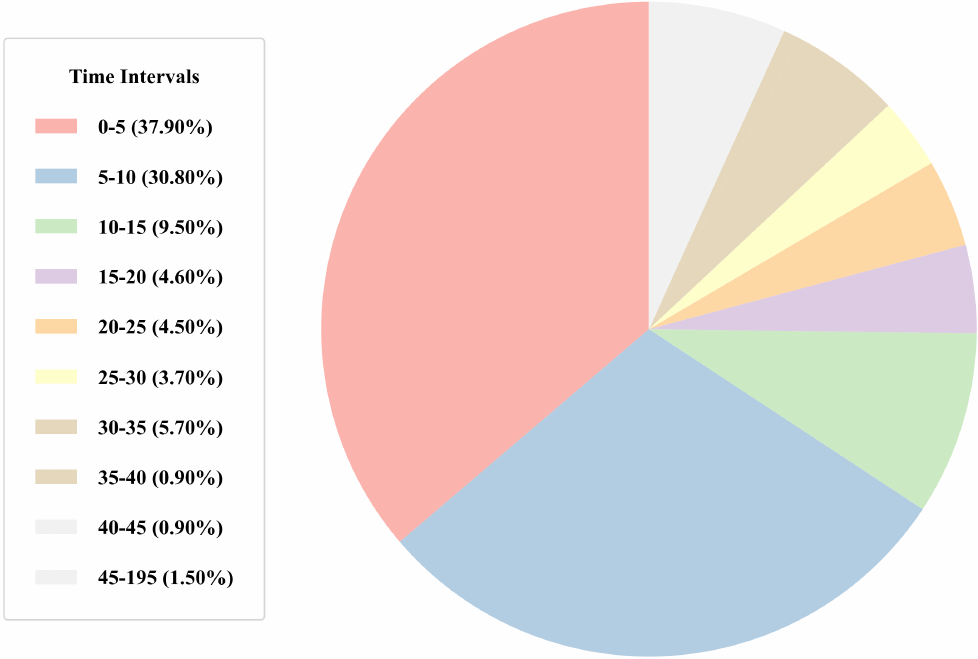}
        \caption{BABEL Length Distribution}
        \label{fig:babel}
    \end{subfigure}
    \caption{Length distribution comparison between \DataName$_{\rm long}$ and BABEL datasets.}
    \label{fig:frame_distribution}
\end{figure*}

Figure \ref{fig:frame_distribution} shows the length distribution comparison between \DataName$_{\rm long}$ and BABEL~\citep{punnakkal2021babel}. 
BABEL labels about 43 hours of mocap sequences from AMASS~\citep{mahmood2019amass} with fine-grained action labels. BABEL exhibits a substantial variation in motion length, containing motion sequences from 5s to over 100s. In BABEL, 37.9\% of motion sequences last 5s or less, which significantly limits its effectiveness for evaluating the long-horizon motion generation capability of T2M models. In contrast, \DataName$_{\rm long}$ contains only 0.33\% of motions within 5s. Furthermore, \DataName$_{\rm long}$ contains 20 times motion sequences than BABEL. As a result, even intervals with relatively low proportions in \DataName$_{\rm long}$ may contain a larger number of motions compared to BABEL. For instance, motions lasting from 35s to 40s only constitute 0.76\% in \DataName$_{\rm long}$, yet \DataName$_{\rm long}$ contains 1,454 motion sequences from 35s to 40s. Meanwhile, although the same interval accounts for a higher proportion (0.9\%) in BABEL, it represents merely 89 motions.

\subsection{Second-wise Text Annotation}
\label{sec:Second-wise_Text_Annotation}
Previous works~\citep{wang2024quo,cao2025being} typically annotate motion sequences by directly feeding corresponding videos into Vision-Language Models (VLMs) to generate coarse textual descriptions. While this approach offers efficiency, it suffers from a critical limitation: motion sequences extracted from web videos often comprise complex and continuous motion clips. When VLMs are applied in an end-to-end manner to entire video clips, they tend to overlook fine-grained and crucial motion details. Such omissions impact the quality and utility of annotated texts, particularly for applications requiring high temporal precision or detailed kinematic analysis.
\begin{figure*}[h]
\centering
    \includegraphics[width=1.0\linewidth]{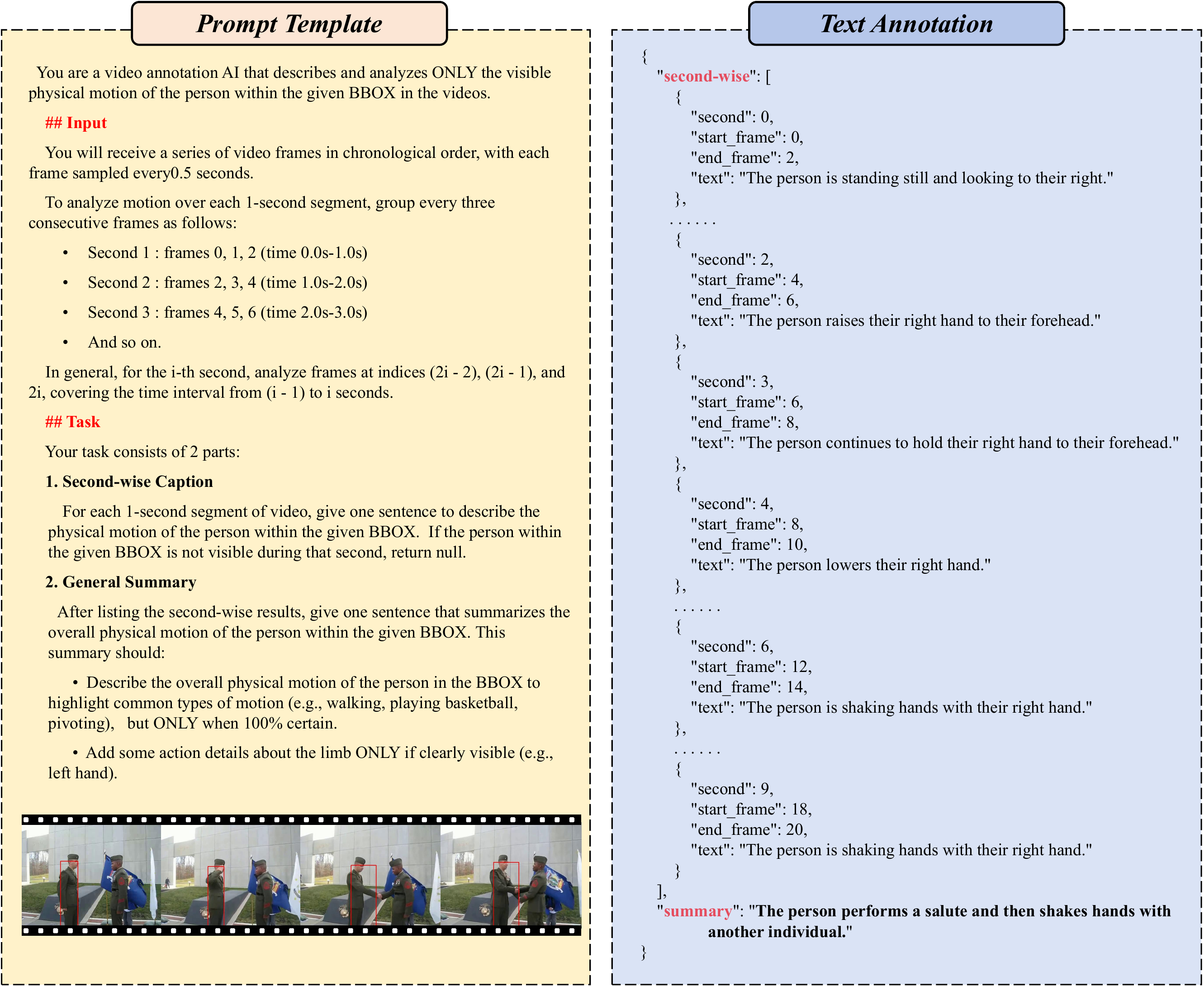}
\caption{Prompt template for generating second-wise text annotations utilizing Gemini-2.5.}
\label{fig:app_video_caption}
\end{figure*}

In this work, we design a second-wise annotation scheme as shown in Figure \ref{fig:app_video_caption}. The annotation task mainly contains second-wise captions and a general summary task. The annotation process begins by uniformly extracting video frames every 0.5s. Each second video frames are first annotated individually with second-wise descriptions. These second-wise captions are then summarized to form a precise caption for the entire video clip. In the annotation process, we deliberately exclude any descriptions of backgrounds, facial expressions, clothes, and other attributes that are irrelevant to human motion. We present annotated examples in Figure \ref{fig:app_vis_examples} to illustrate the precise alignment between text and motion.

\section{Additional Details of 2D-PRQ}
\label{sec:split_part}
In this work, we propose 2D-PRQ, a tokenizer that divides the joints of the whole body into 5 parts, including:
\begin{itemize}
\item Left Hand: spine$_1$, spine$_2$, spine$_3$, left collar, left shoulder, left elbow, left wrist
\item Right Hand: spine$_1$, spine$_2$, spine$_3$, right collar, right shoulder, right elbow, right wrist
\item Left Leg: spine$_1$, spine$_2$, spine$_3$, left hip, left knee, left ankle, left foot
\item Right Leg: spine$_1$, spine$_2$, spine$_3$, right hip, right knee, right ankle, right foot
\item Torso: spine$_1$, spine$_2$, spine$_3$, neck, left collar, right collar, head
\end{itemize}

The pelvis, spine$_1$, spine$_2$, and spine$_3$ are shared across all parts, as they remain relatively stable during human motion. 
Each joint is represented by relative 6D rotations and redundant 3D positions, resulting in a dimensionality of 63+8 per part, including 4D root node and 4D foot contact information. 
When aggregating part features into motion features, we average the shared joints.

\section{Evaluation Metrics}
\label{sec:Evaluation_Metrics}
\textbf{Text-to-motion.} We adapt R-precision, MMDist, and FID to evaluate T2M model follow ~\citet{guo2022generating}. Each metric is illustrated as follows:

\begin{itemize}
    \item \textbf{R-precision}: The retrieval metric is designed to evaluate the semantic consistency between text and generated motion. The R-precision is computed as the accuracy of its ground-truth text description being ranked Top-1 when retrieved by the generated motion from a text pool. Following ~\citet{guo2022generating}, we set the size of the description pool to 32.
    \item \textbf{MMDist:} MultiModel Distance is computed as the average Euclidean distance between motion feature and corresponding text feature.
    \item \textbf{FID}: Frechet Inception Distance is designed to measure the similarity between the distribution of generated motions and ground-truth motion in the feature space. It is computed as the Fréchet distance between the feature distributions of the generated motion and ground-truth motion.
\end{itemize}

\noindent\textbf{Motion Reconstruction.} We adapt FID and MPJPE to evaluate motion tokenizers on the motion reconstruction task.

\begin{itemize}
    \item \textbf{FID}: Similar to T2M, Frechet Inception Distance for motion reconstruction is computed as the Fréchet distance between the feature distributions of reconstruction motion and ground-truth motion.
    \item \textbf{MPJPE}: The metric is computed by averaging the L2 distances between all joints of reconstruction motion and ground-truth motion across all frames.
\end{itemize}

\section{Visualization Examples}
\label{sec:visualiztion_examples}
\begin{figure*}[h]
\centering
    \includegraphics[width=0.62\linewidth]{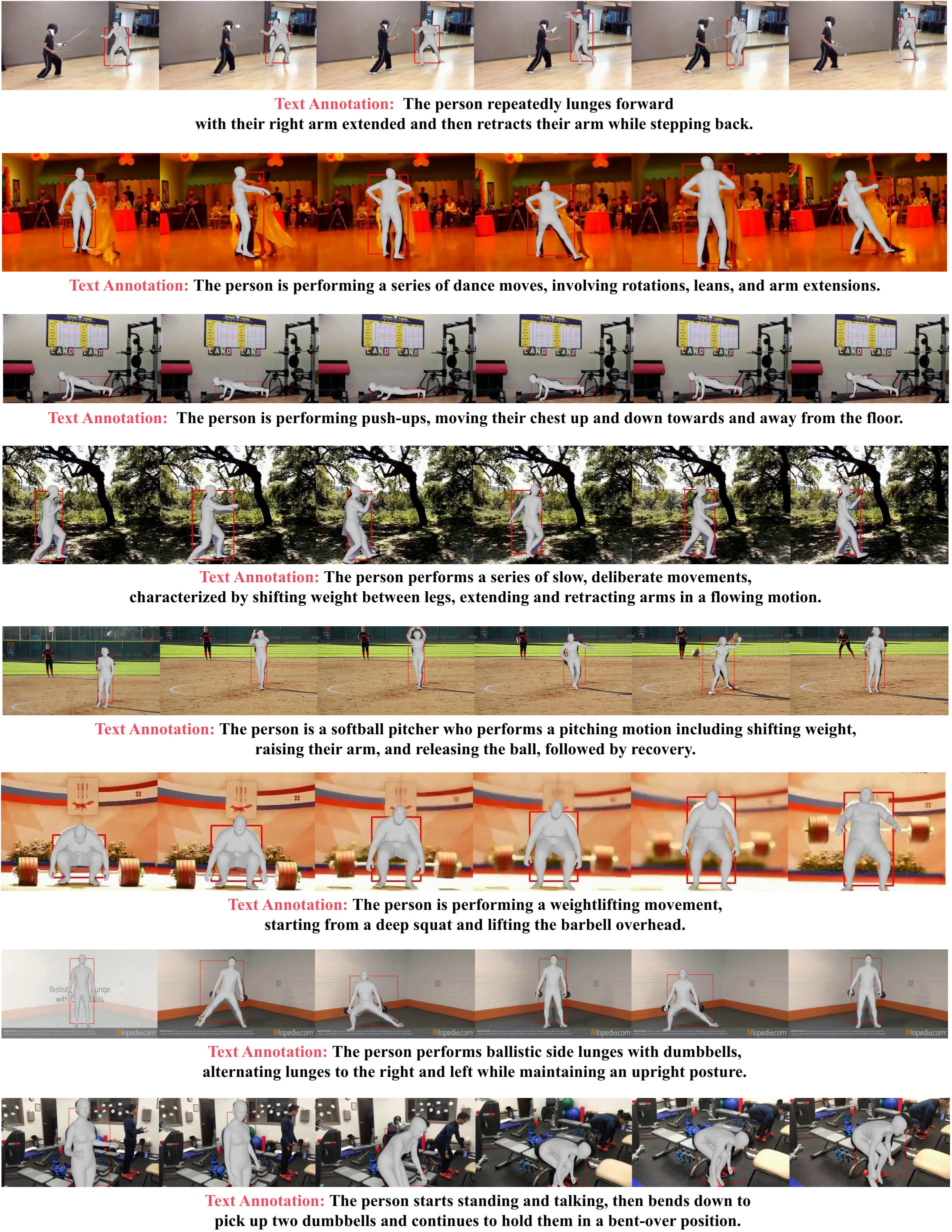}
\caption{Visualization examples of \DataName, each example is annotated with precise text.}
\label{fig:app_vis_examples}
\end{figure*}

We provide visualization examples of \DataName in Figure \ref{fig:app_vis_examples}. Visualization examples demonstrate that \DataName encompasses a diverse range of motion patterns and exhibits strong text-motion alignment, providing a high-quality data foundation for building large motion models.

\clearpage

\end{document}